\documentclass[letterpaper]{article} % DO NOT CHANGE THIS
\usepackage{aaai24}  % DO NOT CHANGE THIS
\usepackage{times}  % DO NOT CHANGE THIS
\usepackage{helvet}  % DO NOT CHANGE THIS
\usepackage{courier}  % DO NOT CHANGE THIS
\usepackage[hyphens]{url}  % DO NOT CHANGE THIS
\usepackage{graphicx} % DO NOT CHANGE THIS
\usepackage{natbib}  % DO NOT CHANGE THIS AND DO NOT ADD ANY OPTIONS TO IT
\usepackage{caption} % DO NOT CHANGE THIS AND DO NOT ADD ANY OPTIONS TO IT
\usepackage{algorithm}
\usepackage{algorithmic}
\usepackage{amssymb}

\usepackage{newfloat}
\usepackage{listings}
\DeclareCaptionStyle{ruled}{labelfont=normalfont,labelsep=colon,strut=off} % DO NOT CHANGE THIS
\floatstyle{ruled}
\newfloat{listing}{tb}{lst}{}
\floatname{listing}{Listing}
\title{Structured Probabilistic Coding}
\author{Dou Hu\textsuperscript{\rm 1,\rm 2}, Lingwei Wei\textsuperscript{\rm 1}, Yaxin Liu\textsuperscript{\rm 1,\rm 2}, Wei Zhou\textsuperscript{\rm 1}\thanks{Corresponding author.}, Songlin Hu\textsuperscript{{\rm 1},{\rm 2}}\footnotemark[1] 
}

\affiliations{
       \textsuperscript{\rm 1} Institute of Information Engineering, Chinese Academy of Sciences \\
       \textsuperscript{\rm 2} School of Cyber Security, University of Chinese Academy of Sciences  \\
       \{hudou, weilingwei, liuyaxin, zhouwei, husonglin\}@iie.ac.cn \\
}

\usepackage{multirow}
\usepackage{amsmath}
\usepackage{amsfonts}
\usepackage{makecell}
\begin{document}

\maketitle 
\begin{abstract}
This paper presents a new supervised representation learning framework, namely structured probabilistic coding (SPC), to learn compact and informative representations from input related to the target task. SPC is an encoder-only probabilistic coding technology with a structured regularization from the target space. It can enhance the generalization ability of pre-trained language models for better language understanding. Specifically, our probabilistic coding simultaneously performs information encoding and task prediction in one module to more fully utilize the effective information from input data. It uses variational inference in the output space to reduce randomness and uncertainty. Besides, to better control the learning process of probabilistic representations, a structured regularization is proposed to promote uniformity across classes in the latent space. With the regularization term, SPC can preserve the Gaussian structure of the latent code and achieve better coverage of the hidden space with class uniformly. Experimental results on 12 natural language understanding tasks demonstrate that our SPC effectively improves the performance of pre-trained language models for classification and regression. Extensive experiments show that SPC can enhance the generalization capability, robustness to label noise, and clustering quality of output representations.

\end{abstract}

\section{Introduction}

Probabilistic embedding \cite{DBLP:journals/corr/VilnisM14} is a flexible representation learning technology whose goal is to learn the underlying probability distribution of data. In contrast to deterministic embedding \cite{DBLP:conf/iclr/PereyraTCKH17,DBLP:conf/iclr/MiyatoDG17,gunel2020supervised}, which maps each data to a fixed vector representation, probabilistic embedding embraces the notion of learning a probability distribution, mapping each data point to a distribution.
These probabilistic embedding approaches can better describe the uncertainty and complexity of data, handle redundant information, and provide better discriminative representations.
Probabilistic embedding has been applied to various domains such as computer vision \cite{DBLP:conf/iclr/OhMPRSG19,DBLP:conf/iccv/ShiJ19} and natural language processing \cite{DBLP:conf/iclr/MahabadiBH21,DBLP:conf/emnlp/0001HDZJMS22}.

Most probabilistic embedding methods \cite{DBLP:journals/corr/KingmaW13,DBLP:conf/iclr/AlemiFD017, DBLP:conf/iclr/HigginsMPBGBML17,DBLP:journals/entropy/Fischer20,DBLP:conf/cvpr/AnJC23} are (or can be) built upon the information bottleneck (IB) principle \cite{tishby1999information,tishby2015deep}.
The principle aims to find a maximally compressed representation of the input that preserves as much as possible information about the target task, striking a balance between compression and prediction.
These IB-based methods typically involve two parametric modules, i.e., an encoder and a decoder \cite{DBLP:journals/jsait/GoldfeldP20}. 
Usually, the encoder maps the input to a probabilistic distribution in the latent space, and the decoder maps the probabilistic distribution to the output representations in the target task space.

However, under the encoder-decoder architecture, the process of mapping input data to probability distributions by the encoder may lose some task-related information, which is essential for the decoder during the learning process. This is because probability distributions inherently contain randomness and uncertainty, which may be irrelevant to the task and interfere with the task prediction process of the decoder.
To avoid this, we propose an encoder-only embedding technology, \textbf{probabilistic coding}, that integrates probabilistic encoding and task prediction into one module.
By using variational inference in the output space, we can better control and utilize randomness and uncertainty of data. 
The learned compact representations can fully capture the underlying structure of data, and preserve the effective information from input related to target task.
This helps improve model generalization performance, especially when facing limited data or noisy labels.

Besides, although probabilistic embedding methods can capture data uncertainty and complexity, they are restricted to limited or biased data, which cannot fully represent the true distribution of the target task. In the process of mapping input data to probability distributions in the latent space by the encoder, some task-related important information may be missing to some extent.
The insufficient information lead to poor task performance on new data and inadequate model generalization.
To improve task prediction ability of the latent representations, we leverage the structured information of target task space to constrain the learning process of the probability distribution in latent space. 
Under the framework of probabilistic coding, the \textbf{structured regularization} of latent space can help the model learn more informative representations related to the target task, thereby improving the model's prediction accuracy on new data.

In this paper, we present a new supervised representation learning framework, \textbf{structured probabilistic coding} (SPC), an encoder-only probabilistic coding technology with a structured regularization from the target label space. By extracting compact and informative representations from input related to the target task, SPC can enhance the generalization ability of pre-trained language models for better language understanding. 
Specifically, the probabilistic coding technique performs variational approximation to encode the input into stochastic output representations under Gaussian distribution spaces, while minimizing the conditional entropy of the target label given the representations.
Besides, the structure information of target task space is introduced to constrain the probability distribution of latent space. 
The structured regularization encourages class-level uniformity within the latent space under the multivariate Gaussian distribution, making the distribution better reflect task-related information, which is beneficial for task prediction.
Under the probabilistic coding framework with the regularization term, SPC can maintain the Gaussian structure of the latent code while achieving the best possible coverage of the hidden space with uniformity across classes.

We conduct experiments on 12 natural language understanding tasks, including 10 classification tasks such as emoji prediction, hate speech detection, irony detection, offensive language detection, sentiment analysis, stance detection, emotion detection from different domains, as well as 2 regression tasks including semantic similarity prediction and plausible clarifications ranking. The results demonstrate that our SPC effectively improves the performance of pre-trained language models for classification and regression tasks. For instance, with the RoBERTa backbone, SPC improves average performance by \textbf{+3.9\%} and \textbf{+1.7\%} for classification and regression tasks compared to CE/MSE. Our SPC framework consistently achieves the best average performance compared to other methods, including deterministic and probabilistic approaches, under different backbone models such as BERT and RoBERTa. Extensive experiments show that SPC can enhance the model generalization capability including out-of-distribution and data-constrained scenarios, robustness to label noise, and clustering quality of output representations.

The main contributions are as follows:    
    1) We propose an encoder-only probabilistic coding method that integrates probabilistic encoding and task prediction into one module. It maximally preserves the effective information from input related to target task.
    2) We design a structured regularization term to promote class-level uniformity in the latent space for better task prediction ability of probabilistic embedding.
    3)  We present a supervised representation learning framework named SPC, to learn compact and informative representations from input related to the target task. It can enhance the generalization ability of pre-trained language models for better language understanding.
    4) Experiments on 12 benchmarks show that SPC achieves state-of-the-art performance on classification and regression tasks. Extensive experiments reveal that SPC can enhance the generalization capability, robustness to label noise, and clustering quality of output representations.\footnote{The code is available at \url{https://github.com/zerohd4869/SPC}}

\begin{figure}[t]
\centering
\includegraphics[width=0.82\columnwidth]{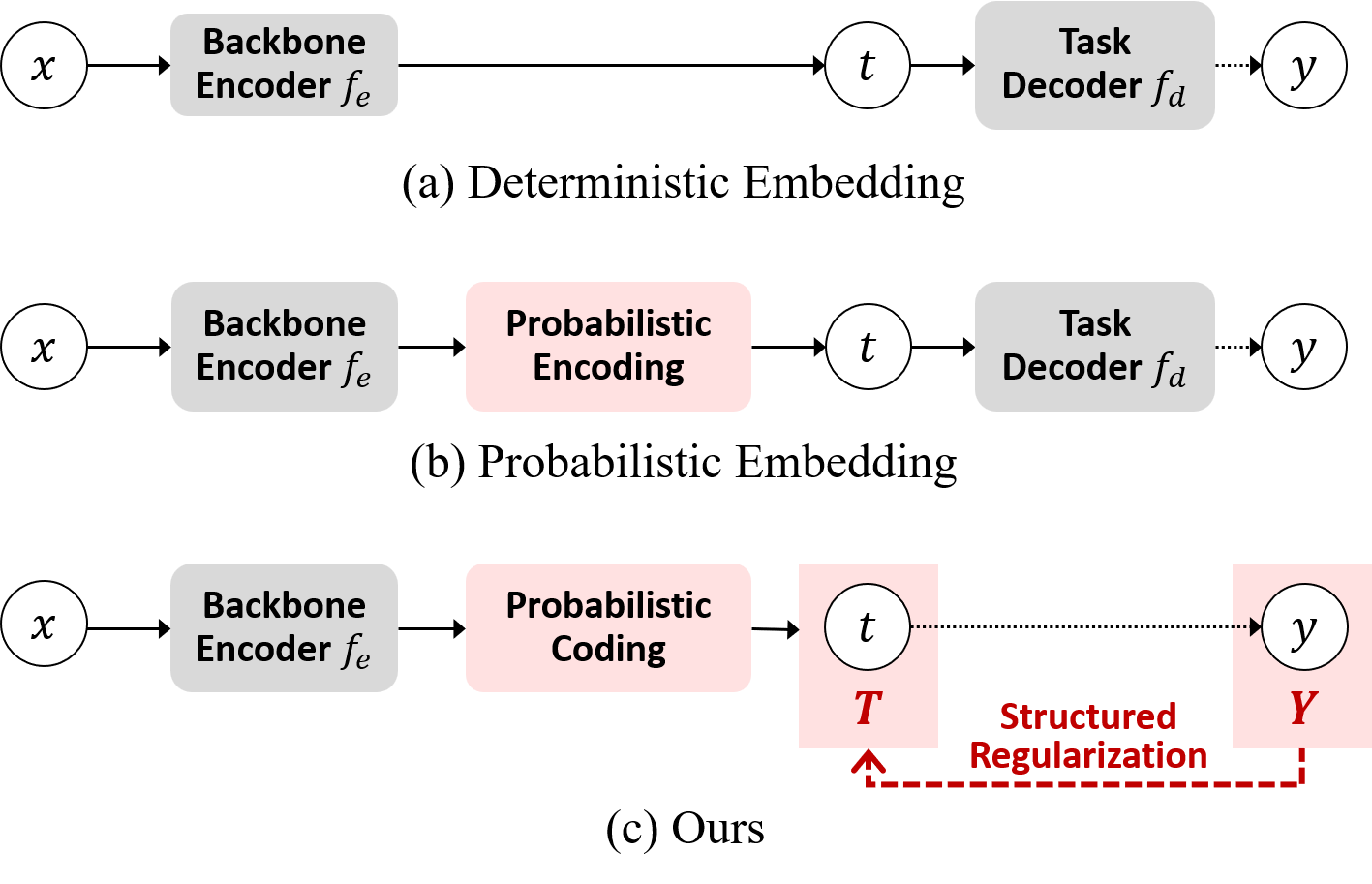}  
\caption{Comparison of our SPC with existing deterministic embedding and probabilistic embedding methods.}
\label{fig1}
\end{figure}

\section{Methodology}
In this section, we present a supervised representation learning framework, \textbf{structured probabilistic coding} (SPC), to learn compact and informative representations from input related to the downstream task. 
As shown in Figure~\ref{fig1}(c), SPC is an encoder-only probabilistic coding technology with a structured regularization from the target label space.

\paragraph{Probabilistic Coding}
The probabilistic coding technology integrates probabilistic encoding and task prediction into one module. Different from existing probabilistic embeddings applying encoder-decoder architecture, our encoder-only paradigm can effectively maintain task-related features and avoid negative effects caused by the randomness and uncertainty of probabilistic encoding.

Under the Markov chain constraint $Y \rightarrow X \rightarrow T$, we have $p(t|x,y)=p(t|x)$.
we aim to minimize the mutual information between the input $X$ and the latent representation $T$, as well as maximize the information between the representations $T$ and the target label $Y$.
Specifically, we employ a variational approximation to encode each input $x$ into a Gaussian distribution representation $t$ in the output space $\mathcal{Y}$, i.e., $T \in \mathbb{R}^{|\mathcal{Y}|}$. 
Additionally, we maximize the lower bound of $I(T;Y)$ 
by estimating the conditional entropy of the target label $Y$ given the representations $T$.
The objective of probabilistic coding can be: 
\begin{equation}
% \resizebox{0.89\linewidth}{!}{$
\small
\begin{split}
      \mathcal{L}_{PC} = \mathbb{E}_{t \sim p_{\theta}(t|x)} [-\log q(y|t)] 
     + \beta  KL(p_{\theta}(t|x); r(t)),
\end{split}
% $}
\end{equation}
where $q(y|t)$ is a non-parametric operation, 
e.g., softmax function for classification.
$r(t)$ is an estimate of the prior $p(t)$ of $t$.
$p_{\theta}(t|x)$ 
is a variational estimate of the posterior probability of $t$ and is learned by the stochastic encoder $\theta$.
$KL(\cdot)$ denotes the analytic KL-divergence term, serving as the regularization that forces the posterior probability of $t$ to approximately converge to the prior $p(t)$.
$\beta > 0$ is a hyperparameter which controls the trade-off between the sufficiency of $t$ for predicting $y$ and the compression of $t$ from $x$.

Let the prior $p(t)$ be the isotropic Gaussian distribution.
And let the variational approximate posterior $p_{\theta}(t|x)$ be
a multivariate Gaussian with a diagonal covariance structure,
i.e., $p_{\theta}(t|x) = \mathcal{N}(t;{\mu}(x), \Sigma(x))$, where $\mu$ and $\Sigma$ represent the mean and diagonal covariance.
Both of their parameters are input-dependent and predicted by an MLP (a fully-connected neural network with a single hidden layer), respectively.
As the sampling of $t$ is a stochastic process, we apply the re-parameterization trick \cite{DBLP:journals/corr/KingmaW13} to ensure unbiased gradients for the model.

In existing IB-based methods \cite{DBLP:conf/iclr/AlemiFD017,DBLP:journals/entropy/Fischer20,DBLP:conf/cvpr/AnJC23} with an encoder-decoder architecture, their decoder can be a parametric approximation $q_{\phi}$ of $p(y|t)$. That is, the compressed representation $t$ can be sampled from the distribution $p_{\theta}(t|x)$, meaning that a specific pattern of noise is added to the input of $q_{\phi}(y|t)$. The noise could diminish the information conveyed by $t$ and potentially cause a loss of task-related information, which is crucial for the decoder $\phi$ during the learning process.
Different from them, our probabilistic coding integrates probabilistic encoding and task prediction into one encoder module with the network $\theta$, and applies a non-parametric operation to obtain prediction outputs. It effectively avoids negative impacts caused by the randomness and uncertainty of probabilistic encoding.

\paragraph{Structured Regularization}
As mentioned above, the Markov assumption restricts that the representation $T$ cannot depend directly on the target task $Y$.
This means that the learning of $t$ does not fully utilize information of label space. 
Accordingly, the learned representation cannot sufficiently represent the true distribution of the target task, leading to poor generalization ability when learning from the limited or biased data.
To enhance the prediction ability of the latent representation, we design a new structured regularization related to the target task space.
It can constrain the learning process of probability distributions in the latent space.

Specifically, we add an additional regularization term about the latent distribution to the objective function that maximizes the prior entropy of $T$ on the label space:
\begin{equation}
\max H(T), \ \text{where} \ T\in\mathbb{R}^{|\mathcal{Y}|}. 
\end{equation}
and 
\begin{equation}
% \resizebox{0.83\linewidth}{!}{$
\small
\begin{split}
H(T) &= -\frac{1}{|\mathcal{X}|}\sum_{i=1}^{|\mathcal{X}|}\sum_{j=1}^{|\mathcal{Y}|}p_{i,j} \log p_{i,j} \\
\end{split}
% $}
\end{equation}
where
$|\mathcal{X}|$ is the number of the expected data.
$|\mathcal{Y}|$ is the number of classes. $p_{i,j}$ represents the probability of sample $i$ belonging to class $j$ in the latent space, which is computed by the output of $p_{\theta}$, i.e., the encoder of probabilistic embedding.

In the implementation, the Jensen's inequality is applied to estimate the upper bound of $H(T)$. Given each sampled batch data, we have:
\begin{equation}
% \resizebox{0.89\linewidth}{!}{$
    \small
\begin{split}
H(T_b) \leq \hat{H}(T_b) & \approx - \sum_{j=1}^{|\mathcal{Y}|} \left ( \frac { 1 } { B } \sum _ { k=1 } ^ { B } p_ { k,j } \right ) \log \left ( \frac { 1 } { B } \sum _ { l=1 } ^ { B } p_ { l,j } \right ) \\ 
   & \approx - \sum_{j=1}^{|\mathcal { Y } | } \overline { p_ { \cdot ,j } } \log \left ( \overline { p_ { \cdot ,j } } \right ) = \mathcal{L}_b,
    \end{split}
    % $}
\end{equation}
where $B$ is the batch size. $\overline{p_{\cdot,j}} = \frac{1}{N}\sum_{i=1}^{N} p_{i,j}, j \in \{1,...,|\mathcal{Y}|\}$ represents the average predicted probability of the $j$-th target label variable.
The Monte Carlo method are used to estimate the expected value of $H(T)$ via sampling batch data and computing the batch entropy $\mathcal{L}_b$, which measures the uncertainty or diversity of the predicted probability distribution over the label space. 
The regularization term encourages uniformity across different classes in the latent space. 
This approach allows for a balanced learning process across different labels, preventing the model from excessively emphasizing certain prevalent features within the training data that might not accurately represent the true data distribution.

\paragraph{Structured Probabilistic Coding}
We incorporate the structured regularization from the target label space into the probabilistic coding framework, named Structured Probabilistic Coding (SPC).
The total objective of SPC can be:
\begin{equation}
% \resizebox{0.89\linewidth}{!}{$
\small
\begin{split}
\mathcal{L}_{SPC} 
&= \mathcal{L}_{PC} - \gamma  H(T)   
 \gtrapprox \mathcal{L}_{PC} - \gamma  \mathcal{L}_b\\ 
&= \mathbb{E}_{t \sim p_{\theta}(t|x)} [-\log q(y|t)] 
     + \beta  KL(p_{\theta}(t|x); r(t)) \\ 
     & \quad   + \gamma  \sum_{j=1}^{|\mathcal { Y } | } \overline { p_ { \cdot ,j } } \log \left ( \overline { p_ { \cdot ,j } } \right ) , 
\end{split}
% $}
\end{equation}
where $\gamma >0$ is a hyperparameter controlling the strength of the regularization.
The first two terms combine probabilistic encoding and task prediction into one encoder module with the network $\theta$. 
The last regularization term promotes class-level uniformity in the latent space under the multivariate Gaussian distribution. 
Totally, the goal of SPC is to maintain the Gaussian structure of the latent code, as well as achieve the best possible coverage of the hidden space with uniformity across classes.

\paragraph{Applications for Downstream Tasks}
We apply the SPC framework to enhance the generalization ability of pre-trained language models for various natural language understanding (NLU) tasks.
Due to the ability of learning informative and compact representations, the proposed SPC framework is well-suited for classification and regression tasks. 
For classification tasks, the lower bound of $I(T;Y)$ can amount to a classic cross-entropy loss \cite{achille2018information,amjad2019learning}.
Similarly, for regression tasks, the lower bound of $I(T;Y)$ can be equivalent to a classic mean squared error loss.

\section{Experiments}

\subsection{Experimental Setups}

\paragraph{Datasets and Downstream Tasks}
We conduct experiments on various classification and regression tasks.
Concretely, following \citet{DBLP:conf/emnlp/BarbieriCAN20}, we experiment on 7 classification tasks about tweet analysis on social media, i.e., 
\textbf{EmojiEval} \cite{DBLP:conf/semeval/BarbieriCRABBPS18},
\textbf{EmotionEval} \cite{DBLP:conf/semeval/MohammadBSK18},
\textbf{HatEval} \cite{DBLP:conf/semeval/BasileBFNPPRS19}, 
\textbf{IronyEval} \cite{DBLP:conf/semeval/HeeLH18},
\textbf{OffensEval} \cite{DBLP:conf/semeval/ZampieriMNRFK19},
\textbf{SentiEval} \cite{DBLP:conf/semeval/RosenthalFN17},
and \textbf{StanceEval} \cite{DBLP:conf/semeval/MohammadKSZC16}.
To better evaluate the generalization of the method for cross-domain scenes, we also experiment on 3 emotion-related datasets from different domains, i.e., \textbf{ISEAR} \cite{scherer1994evidence},
\textbf{MELD} \cite{DBLP:conf/acl/PoriaHMNCM19},
and 
\textbf{GoEmotions} \cite{DBLP:conf/acl/DemszkyMKCNR20}. 
Besides, we experiment on 2 regression benchmarks, i.e., \textbf{STS-B} \cite{DBLP:conf/semeval/CerDALS17} and \textbf{CLAIRE} \cite{DBLP:conf/semeval/RothAS22}.
See the appendix for more descriptions of datasets and tasks.

\paragraph{Comparison Methods}
We compare against the 4 universal models (i.e., SVM, FastText, BiLSTM, and GPT-3.5) and 7 representative deep representation learning technologies (i.e., 
CE/MSE, CE+CP, CE/MSE+AT, CE+SCL, VIB, MINE-IB, and MEIB). 
VIB, MINE-IB, and MEIB belong to probabilistic embedding methods, while the others typically belong to deterministic embedding methods.
For these representation learning technologies, we use pre-trained language models, i.e., BERT \cite{DBLP:conf/naacl/DevlinCLT19}, and RoBERTa \cite{DBLP:journals/corr/abs-1907-11692}, as the backbone models for fine-tuning on downstream tasks.
Concretely, we use \textit{bert-base-uncased}\footnote{\url{https://huggingface.co/}\label{code}} and \textit{roberta-base}\textsuperscript{\ref{code}} to initialize BERT and RoBERTa for fine-tuning on downstream tasks, respectively.

\textbf{SVM} \cite{cortes1995support} is a machine learning algorithm with a hinge loss that aims to find the best hyperplane to separate data points into different classes.
\textbf{FastText} \cite{joulin2017bag} is an efficient text classification method with negative log-likelihood loss based on n-gram features and a hierarchical softmax.
\textbf{BiLSTM} is a bidirectional recurrent neural network \cite{hochreiter1997long} that can be used for classification with cross-entropy loss.
\textbf{GPT-3.5}\footnote{\url{https://chat.openai.com}} is an enhanced generative pre-trained transformer model based on text-davinci-003, optimized for chatting.\footnote{We present the zero-shot results of the GPT-3.5-turbo snapshot from June 13th 2023 based on specific inputs, including task descriptions, task instructions, and evaluation texts.}

\textbf{CE/MSE} means a fine-tuned baseline with a cross-entropy (CE) loss for classification tasks or a mean squared error (MSE) loss for regression tasks.
\textbf{CE+CP} \cite{DBLP:conf/iclr/PereyraTCKH17} is an entropy regularization method that fits a deterministic network by optimizing an objective that combines the CE loss with a confidence penalty term. 
\textbf{CE/MSE+AT} \citep{DBLP:conf/iclr/MiyatoDG17} uses CE/MSE with classical adversarial training.
\textbf{CE+SCL} \cite{gunel2020supervised} combines CE and supervised contrastive learning (SCL) \cite{khosla2020supervised}.
SCL allows for multiple positives per anchor, thus adapting contrastive learning to the fully supervised setting.
\textbf{VIB} \cite{DBLP:conf/iclr/AlemiFD017,DBLP:conf/iclr/MahabadiBH21}  is an efficient variational estimation method of the information bottleneck (IB) principle \cite{tishby2015deep}. 
\textbf{MINE-IB} \cite{DBLP:conf/icml/BelghaziBROBHC18}
is an neural estimation method of the IB principle with a continuous setting.
\textbf{MEIB}  
\cite{DBLP:conf/cvpr/AnJC23} is a variational approach to stochastic embedding in which maximum conditional entropy acts as the bottleneck.
MEIB encourages obvious inputs that can be easily classified to take broader embedding areas by assigning larger entropy.

\paragraph{Evaluation Metrics} \label{sec:eval}
We use the same evaluation metric from the original tasks.
For the evaluation on classification tasks, the macro-averaged F1 over all classes is applied in most cases. There are three exceptions: stance (macro-averaged of F1 of favor and against classes), irony (F1 of ironic class), and sentiment analysis (macro-averaged recall). 
Following \citet{DBLP:conf/emnlp/BarbieriCAN20}, 
we report a global metric based on the average of all dataset-specific metrics.
For the evaluation on regression tasks, 
we apply both Pearson and Spearman correlation coefficients. 
Besides, the $t$-test \cite{kim2015t} is used to verify the statistical significance of the differences between the results of our SPC and the best non-SPC method on the current dataset.

\paragraph{Implementation Details}
All experiments are conducted on a single NVIDIA Tesla A100 80GB card. 
The validation sets are used to tune hyperparameters and choose the optimal model.
For each method, we run five random seeds and report the average result of the test sets.
Besides, we conduct experiments using an epoch number of $20$, a total batch size of $128$, and a maximum token length of $128$. The maximum patience for early stopping is set to 5 epochs. 
The network parameters are optimized by using Adamax optimizer \citep{DBLP:journals/corr/KingmaB14} with the learning rate of $5e^{-5}$, the weight decay coefficient of $\{0, 0.01, 0.001\}$.
For SPC, the trade-off parameter $\beta$ and $\gamma$ are searched from $\{0.001, 0.01, 0.1, 1, 10\}$ respectively. 
More experimental details are listed in the Appendix.

\begin{table*}[t]
\centering
\small
\tabcolsep=1.5pt
% \resizebox{0.96\linewidth}{!}{$
\begin{tabular}{l|cccccccccc|c}
\hline
\multicolumn{1}{c|}{\multirow{1}{*}{{Methods}}} 
& \multicolumn{1}{c}{{EmojiEval}} 
& \multicolumn{1}{c}{{EmotionEval}} 
& \multicolumn{1}{c}{{HatEval}} 
& \multicolumn{1}{c}{{IronyEval}} 
& \multicolumn{1}{c}{{OffensEval}} 
& \multicolumn{1}{c}{{SentiEval}} 
& \multicolumn{1}{c}{{StanceEval}} 
& \multicolumn{1}{c}{{ISEAR}} 
& \multicolumn{1}{c}{{MELD}} 
& \multicolumn{1}{c|}{{GoEmotions}} 
& \multicolumn{1}{c}{\textbf{Avg.}} 
\\  
\hline 
\multicolumn{1}{l|}{SVM$^\dag$} & 29.30 & 64.70 & 36.70 & 61.70 & 52.30 & 62.90 & 67.30 & - & - & - & - \\ 
\multicolumn{1}{l|}{FastText$^\dag$} & 25.80 & 65.20 & 50.60 & 63.10 & 73.40 & 62.90 & 65.40 & - & - & - & - \\ 
\multicolumn{1}{l|}{BiLSTM$^\dag$} & 24.70 & 66.00 & 52.60 & 62.80 & 71.70 & 58.30 & 59.40 & - & - & - & - \\ 
\multicolumn{1}{l|}{GPT-3.5} & {6.34}\tiny{$\pm$0.01} & {73.23}\tiny{$\pm$0.18} &  {48.30}\tiny{$\pm$0.11} & \textbf{66.81}\tiny{$\pm$3.26} & {63.71}\tiny{$\pm$0.13}  & {40.40}\tiny{$\pm$3.13}  & {39.45}\tiny{$\pm$0.10}  & {67.22}\tiny{$\pm$0.09}  & {41.46}\tiny{$\pm$0.11}  & {25.21}\tiny{$\pm$0.08}  & {47.21} \\ 
\hline
\multicolumn{10}{l}{\multirow{1}{*}{\textit{BERT backbone}}} \\  \hline 
\multicolumn{1}{l|}{CE} &  
{22.30}\tiny{$\pm$0.60} & {76.05}\tiny{$\pm$1.41} & {44.67}\tiny{$\pm$1.78} & {59.38}\tiny{$\pm$3.01} & {80.16}\tiny{$\pm$1.26} & {70.54}\tiny{$\pm$0.44} & {65.21}\tiny{$\pm$0.71} & {67.17}\tiny{$\pm$0.78} & {39.80}\tiny{$\pm$0.84} & {46.29}\tiny{$\pm$0.79} & 57.16 \\  
\multicolumn{1}{l|}{CE+CP} & 21.91\tiny{$\pm$0.71} & 76.28\tiny{$\pm$1.20} & 45.97\tiny{$\pm$2.93} & 64.06\tiny{$\pm$2.41} & 78.99\tiny{$\pm$1.57} & 70.68\tiny{$\pm$0.31} & 65.83\tiny{$\pm$0.39} & 67.20\tiny{$\pm$0.95} & 39.54\tiny{$\pm$1.61} & 46.39\tiny{$\pm$0.63} & 57.69
\\ 
\multicolumn{1}{l|}{CE+AT} & 
22.93\tiny{$\pm$0.70} & 75.08\tiny{$\pm$1.23} & 46.30\tiny{$\pm$3.61} & 64.23\tiny{$\pm$2.04} & 79.68\tiny{$\pm$1.59} & 70.55\tiny{$\pm$0.57} & 66.46\tiny{$\pm$1.13} & 65.70\tiny{$\pm$0.69} & 39.84\tiny{$\pm$0.38} & 47.37\tiny{$\pm$0.54} &  57.81     \\
\multicolumn{1}{l|}{CE+SCL} & 21.72\tiny{$\pm$0.51} & 75.43\tiny{$\pm$1.37} & 45.86\tiny{$\pm$1.15} & 65.39\tiny{$\pm$2.46} & 80.20\tiny{$\pm$0.56} & 70.70\tiny{$\pm$0.79} & 65.34\tiny{$\pm$0.60}  & 67.54\tiny{$\pm$0.64} & 40.00\tiny{$\pm$1.96} & 46.50\tiny{$\pm$0.46} & 57.87  \\  
\multicolumn{1}{l|}{VIB} &  21.31\tiny{$\pm$0.62} & 77.37\tiny{$\pm$0.71} & 45.99\tiny{$\pm$1.93} & 63.82\tiny{$\pm$1.00} & 80.37\tiny{$\pm$1.11} & 70.39\tiny{$\pm$0.31} & 65.43\tiny{$\pm$0.60} & 67.24\tiny{$\pm$0.57} & 38.52\tiny{$\pm$0.51} & 45.89\tiny{$\pm$1.10} & 57.63
\\
\multicolumn{1}{l|}{MINE-IB} & 
21.29\tiny{$\pm$0.31} & 76.60\tiny{$\pm$0.41} & 47.64\tiny{$\pm$2.11} & 65.86\tiny{$\pm$2.57} & 78.67\tiny{$\pm$2.28} & 69.85\tiny{$\pm$0.54} & 65.35\tiny{$\pm$0.88} & 67.62\tiny{$\pm$0.40} & 41.23\tiny{$\pm$0.67} & 46.87\tiny{$\pm$0.42} & 58.10
\\ 
\multicolumn{1}{l|}{MEIB} & 21.87\tiny{$\pm$0.73} & 76.70\tiny{$\pm$0.82} & 48.27\tiny{$\pm$1.72} & 65.87\tiny{$\pm$2.14} & 80.49\tiny{$\pm$0.81} & 
70.55\tiny{$\pm$0.57} & 65.59\tiny{$\pm$1.58} & 67.44\tiny{$\pm$0.50} & 39.30\tiny{$\pm$0.61} & 46.26\tiny{$\pm$0.81} & 58.23  
\\
\multicolumn{1}{l|}{\textbf{SPC}} &  
{24.19}\tiny{$\pm$1.55} & {77.15}\tiny{$\pm$0.73} & {57.48}\tiny{$\pm$2.99} & 
{65.85}\tiny{$\pm$1.07} 
& {80.65}\tiny{$\pm$0.78} & {70.74}\tiny{$\pm$0.12} & {67.17}\tiny{$\pm$1.08} & {68.94}\tiny{$\pm$0.35} & {42.68}\tiny{$\pm$0.94} & {47.62}\tiny{$\pm$1.38} & {60.25} \\ 
\hline
\multicolumn{10}{l}{\multirow{1}{*}{\textit{RoBERTa backbone}}}  \\ \hline 
CE & 
30.25\tiny{$\pm$1.32} & 77.41\tiny{$\pm$1.33} & 45.49\tiny{$\pm$4.70} & 57.99\tiny{$\pm$4.96} & 78.74\tiny{$\pm$2.20} & 71.80\tiny{$\pm$0.93} & 66.78\tiny{$\pm$1.34} & 70.00\tiny{$\pm$0.45} & 39.23\tiny{$\pm$0.41} & 46.64\tiny{$\pm$1.15} & 58.43 \\  
CE+CP & 
31.12\tiny{$\pm$0.84} & 77.54\tiny{$\pm$0.70} & 48.59\tiny{$\pm$3.28} & 58.75\tiny{$\pm$6.19} & 79.50\tiny{$\pm$0.98} & 72.82\tiny{$\pm$0.29} & 66.89\tiny{$\pm$1.67} & 70.58\tiny{$\pm$0.71} & 40.74\tiny{$\pm$0.89} & 47.98\tiny{$\pm$0.65} & 59.45 \\  
\multicolumn{1}{l|}{CE+AT} & 
32.00\tiny{$\pm$0.93} & 77.30\tiny{$\pm$1.07} & 44.71\tiny{$\pm$4.76} & 60.17\tiny{$\pm$3.17} & 79.81\tiny{$\pm$1.11} & 72.51\tiny{$\pm$0.44} & 67.81\tiny{$\pm$0.95} & 70.97\tiny{$\pm$0.68} & 40.10\tiny{$\pm$0.60} & 47.89\tiny{$\pm$1.21} & 59.33   \\ 
CE+SCL & 
31.09\tiny{$\pm$1.85} & 76.98\tiny{$\pm$2.02} & 49.51\tiny{$\pm$2.86} & 60.71\tiny{$\pm$4.23} & 80.39\tiny{$\pm$0.83} & 
\textbf{73.16}\tiny{$\pm$0.44} 
& 66.73\tiny{$\pm$1.54} & 70.26\tiny{$\pm$0.45} & 40.64\tiny{$\pm$1.02} & 47.87\tiny{$\pm$0.86} & 59.72  \\    
VIB &   
{29.71}\tiny{$\pm$0.79} & {77.99}\tiny{$\pm$0.86} & {49.39}\tiny{$\pm$3.08} & {59.93}\tiny{$\pm$4.57} & {79.63}\tiny{$\pm$0.66} & {72.81}\tiny{$\pm$0.39} & {68.40}\tiny{$\pm$0.52} & {70.74}\tiny{$\pm$0.44} & {38.94}\tiny{$\pm$0.55} & {46.23}\tiny{$\pm$0.18}  & {59.38} 
\\
MINE-IB & 
31.70\tiny{$\pm$0.45} &  {78.79}\tiny{$\pm$0.58} & {46.39}\tiny{$\pm$2.82} & {57.39}\tiny{$\pm$8.27} & 79.76\tiny{$\pm$0.67} & 72.85\tiny{$\pm$0.56} & 67.27\tiny{$\pm$1.00} & 70.15\tiny{$\pm$0.58} & 41.80\tiny{$\pm$2.14} & 48.88\tiny{$\pm$1.04} & 59.50 \\ 
MEIB 
& {29.94}\tiny{$\pm$1.30} 
& 78.73\tiny{$\pm$0.90}  & 49.34\tiny{$\pm$2.42} 
& {60.54}\tiny{$\pm$2.70}  
& 79.68\tiny{$\pm$0.98} & 72.78\tiny{$\pm$0.29} & 67.89\tiny{$\pm$1.70} & 70.86\tiny{$\pm$0.61} 
& 39.00\tiny{$\pm$0.37}  
&
47.18\tiny{$\pm$1.15} & {59.59} \\   
\textbf{SPC} & \textbf{32.54}$^{*}$\tiny{$\pm$0.48} & \textbf{79.01}\tiny{$\pm$0.61} & \textbf{59.80}$^{*}$\tiny{$\pm$1.32} & 65.31\tiny{$\pm$1.91} & \textbf{80.98}\tiny{$\pm$1.36} & {72.96}\tiny{$\pm$0.22} & \textbf{69.02}$^{*}$\tiny{$\pm$0.63} & \textbf{71.01}$^{*}$\tiny{$\pm$0.59} & \textbf{43.99}$^{*}$\tiny{$\pm$0.29} & \textbf{48.92}\tiny{$\pm$1.83} & \textbf{62.35} 
\\
\hline
\end{tabular}
% $}
\caption{Classification evaluation (\%) on 10 benchmark datasets.  
$^\dag$ means the results are from \citet{DBLP:conf/emnlp/BarbieriCAN20}. 
For other methods,
we run five random seeds and report the average result on test sets. BERT and RoBERTa are the backbone models for deep representation learning technologies.
Best results for each dataset are highlighted in bold.  
$^{*}$ represents statistical significance over state-of-the-art scores under the $t$ test ($p < 0.05$).
}
\label{tab:classification}
\end{table*}

\begin{table*}[t]
\centering
\small
\tabcolsep=1pt
% \resizebox{0.96\linewidth}{!}{$
  \begin{tabular}{l|cccccccccc|c}
  \hline
  \multicolumn{1}{c|}{\multirow{1}{*}{{Methods}}} 
  & \multicolumn{1}{c}{{EmojiEval}} 
  & \multicolumn{1}{c}{{EmotionEval}} 
  & \multicolumn{1}{c}{{HatEval}} 
  & \multicolumn{1}{c}{{IronyEval}} 
  & \multicolumn{1}{c}{{OffensEval}} 
  & \multicolumn{1}{c}{{SentiEval}} 
  & \multicolumn{1}{c}{{StanceEval}} 
  & \multicolumn{1}{c}{{ISEAR}} 
  & \multicolumn{1}{c}{{MELD}} 
  & \multicolumn{1}{c|}{{GoEmotions}} 
  & \multicolumn{1}{c}{\textbf{Avg.}} 
  \\  
  \hline 
\textbf{SPC}   &  \textbf{32.54}\tiny{$\pm$0.48} & \textbf{79.01}\tiny{$\pm$0.61} & \textbf{59.80}\tiny{$\pm$1.32} & \textbf{65.31}\tiny{$\pm$1.91} & \textbf{80.98}\tiny{$\pm$1.36} & \textbf{72.96}\tiny{$\pm$0.22} & \textbf{69.02}\tiny{$\pm$0.63} & \textbf{71.01}\tiny{$\pm$0.59} & \textbf{43.99}\tiny{$\pm$0.29} & \textbf{48.92}\tiny{$\pm$1.83} & \textbf{62.35} 
\\  
\ \ - w/o S. &  
{30.98}\tiny{$\pm$0.89} & {78.60}\tiny{$\pm$0.54}  & {56.64}\tiny{$\pm$8.75}  & {62.12}\tiny{$\pm$6.97}  
& {79.16}\tiny{$\pm$1.28} & {72.23}\tiny{$\pm$0.77}  & 68.90\tiny{$\pm$0.60}  & {70.79}\tiny{$\pm$0.22} & {43.75}\tiny{$\pm$0.67} & {48.84}\tiny{$\pm$1.78}  & 61.20
 \\ 
\ \ - w/o S. \& P. 
& 30.25\tiny{$\pm$1.32} & 77.41\tiny{$\pm$1.33} & 45.49\tiny{$\pm$4.70} & 57.99\tiny{$\pm$4.96} & 78.74\tiny{$\pm$2.20} & 71.80\tiny{$\pm$0.93} & 66.78\tiny{$\pm$1.34} & 70.00\tiny{$\pm$0.45} & 39.23\tiny{$\pm$0.41} & 46.64\tiny{$\pm$1.15} & 58.43 \\
\hline
\end{tabular}
% $}
  \caption{Ablation results (\%) on classification tasks. w/o S. indicates removing the structured regularization. w/o P. refers to removing probabilistic coding.  We experiment with RoBERTa backbone.
  }
  \label{tab:abla}
\end{table*}

\begin{table}[t]
\centering
\small
\tabcolsep=1.5pt
% \resizebox{0.96\linewidth}{!}{$
\begin{tabular}{l|cc|cc|c}
\hline 
\multicolumn{1}{c|}{\multirow{2}{*}{Methods}}    &  \multicolumn{2}{c|}{STS-B} & \multicolumn{2}{c|}{CLAIRE}  & \multirow{2}{*}{\bf Avg.}   \\  
& Spearman & Pearson & Spearman & Pearson  \\ 
\hline
MSE & {88.33}\tiny{$\pm$0.32} & {88.80}\tiny{$\pm$0.36} & 
{50.37}\tiny{$\pm$5.90} & {49.10}\tiny{$\pm$5.74} & 69.15 
\\
MSE+AT & 
{88.40}\tiny{$\pm$0.50} & {89.01}\tiny{$\pm$0.37} & {53.09}\tiny{$\pm$0.64} & {51.87}\tiny{$\pm$0.65} & 70.59
\\
VIB &  {88.45}\tiny{$\pm$0.50} & {89.01}\tiny{$\pm$0.40} & {52.86}\tiny{$\pm$0.88} & {51.66}\tiny{$\pm$0.78} & 70.49 \\ 
MEIB & {88.61}\tiny{$\pm$0.14} & {89.13}\tiny{$\pm$0.17} & {52.85}\tiny{$\pm$0.72} & {51.39}\tiny{$\pm$0.81} & 70.50
 \\ 
\textbf{SPC} &    \textbf{88.71}\tiny{$\pm$0.19} & \textbf{89.31}\tiny{$\pm$0.24} & \textbf{53.11}\tiny{$\pm$0.95} & \textbf{52.21}\tiny{$\pm$0.81} & \textbf{70.84} \\
\hline
\end{tabular}
% $}
\caption{Regression evaluation (\%) on 2 benchmark datasets with RoBERTa backbone.
For each method, we run five random seeds and report the average result on test sets. }
\label{tab:Regression}
\end{table}

\subsection{Overall Results}
\paragraph{Performance on Classification Tasks}
The overall results for 10 classification tasks are summarized in Table~\ref{tab:classification}.
Our SPC consistently obtains the best average performance over comparison methods.
When using BERT and RoBERTa backbones, SPC can enhance average performance by \textbf{+3.1\%} and \textbf{+3.9\%} compared to CE for all classification tasks, respectively.
The results indicate the good generalization ability of our method to unseen test sets and show the superiority on classification tasks. 
We notice that SPC achieves big improvements for HatEval and IronyEval, i.e., \textbf{+14.3\%} macro-F1 scores and \textbf{+7.3\%} F1 scores of the ironic class, respectively.
In HateEval, there is a topic distribution disparity between the validation set and the test set. Additionally, the IronyEval task requires complex semantic understanding, with subtle differences between ironic and non-ironic texts. These results indicate that our SPC has a good generalization capability on the above specific scenarios, i.e., topic shifts and subtle semantic labels.

\paragraph{Performance on Regression Tasks}
Table~\ref{tab:Regression} presents the overall results of comparison methods in terms of Spearman and Pearson correlation coefficients for two regression tasks.
SPC obtains better regression results on both datasets. Besides, when using RoBERTa backbone, compared to MSE, SPC achieves \textbf{+1.7\%} absolute improvements in terms of the average performance. 
This demonstrates the superiority and generalization of SPC to unseen test sets on regression tasks.

\subsection{Ablation Study}
We conduct ablation studies by removing the structured regularization (w/o S.) and probabilistic coding (w/o P.).
For classification, Table~\ref{tab:abla} shows the ablation results on all tasks.
When removing the structured regularization, the ablated model obtains inferior performance in terms of all classification metrics.
When further removing probabilistic coding, the results decline significantly. It reveals the effectiveness of structured regularization and probabilistic coding. 
For regression, since the label space is a one-dimensional real-valued score, our SPC is degraded to probabilistic coding. 
The ablation removing probabilistic coding is equivalent to MSE.
From Table~\ref{tab:Regression}, the average performance declines 1.7\%  on regression metrics, which confirms the effectiveness of probabilistic coding for regression.

\subsection{Generalization Evaluation}
We further evaluate the generalization capability of SPC under the following two settings: training with limited data and testing in out-of-distribution (OOD) scenarios.

\begin{figure*}[t]
\centering
\includegraphics[width=0.95\textwidth]{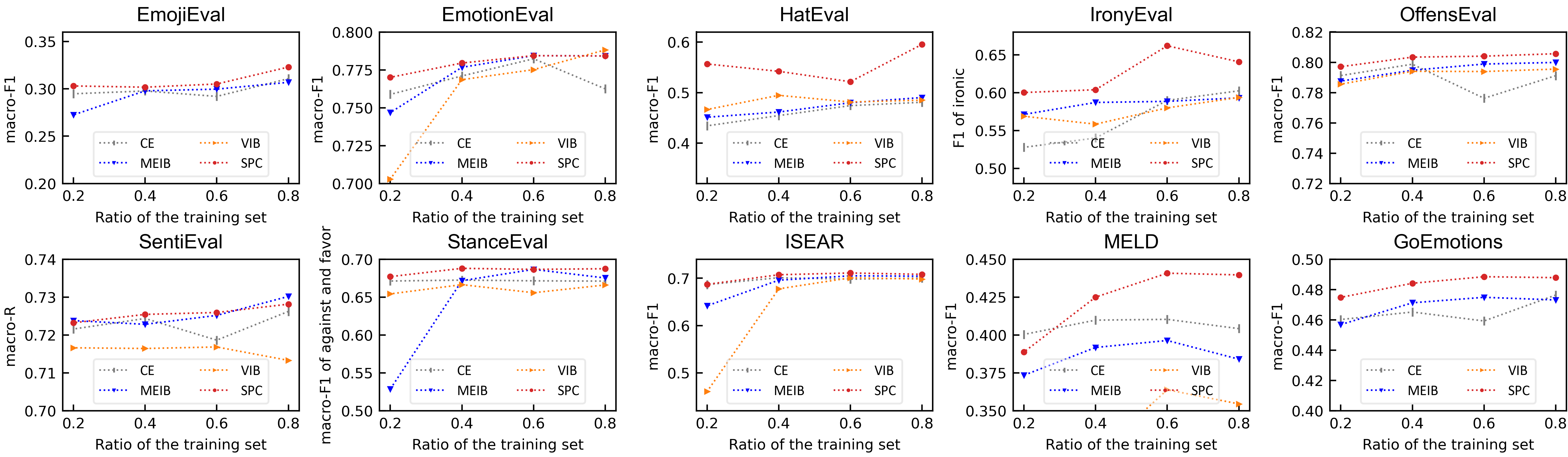}  
\caption{Results of different methods against different sizes of training set with RoBERTa backbone.}
\label{fig:training}
\end{figure*}

\paragraph{Comparison under Different Training Size}
We experiment under different ratios of the training set to evaluate the generalization when training with limited data. 
Specifically, given a predefined ratio (e.g., 20\%) and a random seed, we randomly sample from the original training set. 
We obtain 5 training subsets by independently and repeatedly sampling five times from the original training set with 5 different random seeds.
All methods are trained on these 5 subsets of the training set, and we report the average results on the test set. 
Figure~\ref{fig:training} shows results of CE, VIB, MEIB, and our SPC against different sizes of training set with RoBERTa backbone. 
compared to CE, VIB, and MEIB, SPC achieves superior performance on most datasets against different ratios of the training set.
It indicates that SPC can enhance the generalization ability of pre-trained language models, even when dealing with limited training data.

\begin{table}[t]
\centering
% \resizebox{0.96\linewidth}{!}{$
% \begin{tabular}{l|p{2.6cm}<\centering|p{2.2cm}<\centering|p{2.2cm}<\centering|c}
\small 
\tabcolsep=2pt
\begin{tabular}{l|c|c|c|c}
\hline
\multicolumn{1}{c|}{\multirow{1}{*}{Methods}} 
& \makecell{EmotionEval $\rightarrow$ \\  GoEmotions}
& \makecell{ISEAR $\rightarrow$ \\ GoEmotions}
& \makecell{MELD $\rightarrow$ \\ GoEmotions} 
&  \multirow{1}{*}{Avg.}
\\ 
\hline 
CE & {73.79}\tiny{$\pm$2.57} & {42.99}\tiny{$\pm$2.10} & {30.71}\tiny{$\pm$0.54} & 49.16 \\ 
CE+AT & {72.54}\tiny{$\pm$3.89} & {44.11}\tiny{$\pm$1.44} & {32.05}\tiny{$\pm$1.69}  & 49.57 \\ 
VIB & {74.73}\tiny{$\pm$3.52} & {41.88}\tiny{$\pm$1.65} & {30.50}\tiny{$\pm$1.05} & 49.03 \\ 
MEIB & {75.55}\tiny{$\pm$2.05} & {42.10}\tiny{$\pm$0.61} & {30.11}\tiny{$\pm$1.33} & 49.25 \\ 
\textbf{SPC} & \textbf{77.47}\tiny{$\pm$2.46} & \textbf{44.36}\tiny{$\pm$1.29} & \textbf{33.95}\tiny{$\pm$1.16}  & \textbf{51.93}  
\\
 \hline
\end{tabular}
% $}
\caption{Out-of-distribution evaluation results (\%).  
For instance, ``EmotionEval $\rightarrow$ GoEmotions" refers to training the model on the training set of EmotionEval and making predictions using the test set of GoEmotions.
We experiment with RoBERTa backbone.
We run five random seeds and report the average results on test sets of target domains. Labels that do not appear in the training corpus are not evaluated. 
}
\label{tab:ood-performance}
\end{table}

\paragraph{Evaluation on Out-of-Distribution}
We choose emotion-related benchmarks including EmotionEval, ISEAR, MELD, and GoEmotions, which aim to predict the emotional state but are collected from different domains.
To implement OOD scenarios, we train the model on the original training set from a source domain, select the best model based on the validation set of the source domain, and test on the test set of a target domain. 
To avoid the interference of label mapping bias between different taxonomies, each model is trained on the dataset with coarse-grained taxonomy to predict the label for another dataset with fine-grained taxonomy.
Table~\ref{tab:ood-performance} shows the performance under OOD scenarios.
Our SPC obtains the best results on all OOD settings. The fact exhibits SPC's better generalization capabilities in handling OOD scenarios across different domain shifts. 
It indicates that SPC can better control and utilize randomness and uncertainty of data under the probabilistic coding framework. 
Moreover, the structure regularization from the target task space makes the learned probabilistic representations better reflect task information and generalize the model to new data.

\begin{table*}[!h]
\centering
% \resizebox{0.96\linewidth}{!}{$
\small 
\tabcolsep=1pt
\begin{tabular}{l|c|cccccccccc|c}
\hline
Methods & Noisy & EmojiEval & EmotionEval & HatEval & IronyEval & OffensEval & SentiEval & StanceEval & ISEAR & MELD & GoEmotions & \textbf{Avg.} \\
\hline
CE & 10\% & {30.66}\tiny{$\pm$0.89} & {78.15}\tiny{$\pm$0.88} & {47.06}\tiny{$\pm$5.40} &  {56.90}\tiny{$\pm$4.58} & {79.46}\tiny{$\pm$0.80} & {72.36}\tiny{$\pm$0.74} & {67.39}\tiny{$\pm$1.86} & {70.40}\tiny{$\pm$0.97} & {42.01}\tiny{$\pm$1.94} & {47.85}\tiny{$\pm$1.08} & 59.22  \\ 
VIB & 10\% & {30.74}\tiny{$\pm$0.48} & {77.78}\tiny{$\pm$2.05} & {47.64}\tiny{$\pm$1.57} & {58.66}\tiny{$\pm$10.60} & {79.96}\tiny{$\pm$0.73} & {72.13}\tiny{$\pm$0.54} & {67.54}\tiny{$\pm$1.20}& {70.85}\tiny{$\pm$0.33} & {38.63}\tiny{$\pm$0.89} & {47.30}\tiny{$\pm$1.65} &  59.12  \\ 
MEIB & 10\% & {31.02}\tiny{$\pm$0.47}  & \textbf{78.94}\tiny{$\pm$0.46} & {49.28}\tiny{$\pm$4.58} & {57.21}\tiny{$\pm$8.07} &  \textbf{80.19}\tiny{$\pm$0.83} & {72.09}\tiny{$\pm$0.68} & {68.26}\tiny{$\pm$0.68} & {70.85}\tiny{$\pm$0.38} & {38.67}\tiny{$\pm$0.97} & {46.93}\tiny{$\pm$1.06} & 59.34
\\
\textbf{SPC} &  10\%  & \textbf{32.25}\tiny{$\pm$0.69} & {78.88}\tiny{$\pm$0.47} & \textbf{56.13}\tiny{$\pm$5.36} & \textbf{58.88}\tiny{$\pm$4.94} & 80.14\tiny{$\pm$0.28} & \textbf{72.76}\tiny{$\pm$0.06} & \textbf{68.57}\tiny{$\pm$1.01} & \textbf{71.10}\tiny{$\pm$0.62} & \textbf{43.90}\tiny{$\pm$1.13} & \textbf{49.03}\tiny{$\pm$1.41} & \textbf{61.16} \\
\hline
CE & 20\% & {31.96}\tiny{$\pm$0.88} & {77.01}\tiny{$\pm$1.51} & {49.12}\tiny{$\pm$0.72} & {60.82}\tiny{$\pm$3.56} &{79.54}\tiny{$\pm$1.64}  & {72.06}\tiny{$\pm$0.63} & {68.49}\tiny{$\pm$1.20} & {70.32}\tiny{$\pm$0.26} & {40.16}\tiny{$\pm$1.94} & {47.78}\tiny{$\pm$0.84} &  59.73
\\
VIB & 20\%  & {30.46}\tiny{$\pm$0.59} & \textbf{79.00}\tiny{$\pm$0.49} & {47.91}\tiny{$\pm$2.20} & {60.67}\tiny{$\pm$4.82} & {79.15}\tiny{$\pm$1.22} & {72.26}\tiny{$\pm$0.29} & {66.83}\tiny{$\pm$0.52} & {71.02}\tiny{$\pm$0.25} & {39.33}\tiny{$\pm$1.47} & {47.83}\tiny{$\pm$1.38} & 59.45 \\  
MEIB & 20\% & {30.84}\tiny{$\pm$0.75} & {78.38}\tiny{$\pm$0.88} & {50.02}\tiny{$\pm$5.18} & {55.12}\tiny{$\pm$7.07} & {78.17}\tiny{$\pm$2.55} & {71.63}\tiny{$\pm$1.11} & {68.05}\tiny{$\pm$0.81} & {70.68}\tiny{$\pm$0.38} & {39.09}\tiny{$\pm$0.87} & {47.29}\tiny{$\pm$1.22} & 58.93
\\
\textbf{SPC} & 20\% & \textbf{32.51}\tiny{$\pm$0.83} & 77.97\tiny{$\pm$1.12} & \textbf{55.41}\tiny{$\pm$6.00} & \textbf{66.40}\tiny{$\pm$4.26} & \textbf{80.33}\tiny{$\pm$0.48} & \textbf{72.50}\tiny{$\pm$0.55} & \textbf{68.89}\tiny{$\pm$1.60} & \textbf{71.10}\tiny{$\pm$0.39} & \textbf{43.96}\tiny{$\pm$0.50} & \textbf{50.26}\tiny{$\pm$0.79}
 & \textbf{61.93}
\\
\hline
CE & 30\% & {31.82}\tiny{$\pm$0.75} & {77.61}\tiny{$\pm$0.90} & {50.69}\tiny{$\pm$2.80} & {58.90}\tiny{$\pm$11.45} & {78.11}\tiny{$\pm$2.07} & {70.15}\tiny{$\pm$0.5} & {69.07}\tiny{$\pm$1.07} & {70.74}\tiny{$\pm$0.56} & {40.61}\tiny{$\pm$2.06} & {47.76}\tiny{$\pm$2.29} & 59.55
\\
VIB & 30\% & {30.85}\tiny{$\pm$0.53} & \textbf{78.23}\tiny{$\pm$0.79} & {48.22}\tiny{$\pm$1.97} & {58.81}\tiny{$\pm$8.84} & {79.38}\tiny{$\pm$0.62} & {72.15}\tiny{$\pm$0.52} & {67.59}\tiny{$\pm$0.93} & {70.27}\tiny{$\pm$0.74} & {38.71}\tiny{$\pm$1.19} & {47.16}\tiny{$\pm$1.32} & 59.14 \\ 
MEIB & 30\% & {30.74}\tiny{$\pm$0.87} & {77.99}\tiny{$\pm$0.69} & {49.98}\tiny{$\pm$4.00} & {57.57}\tiny{$\pm$5.19} & {72.53}\tiny{$\pm$5.53} & {71.83}\tiny{$\pm$0.40} & {67.88}\tiny{$\pm$0.68} & {69.86}\tiny{$\pm$1.24} & {39.39}\tiny{$\pm$1.06} & {47.43}\tiny{$\pm$1.52} & 58.52 
 \\
\textbf{SPC} & 30\% & \textbf{32.27}\tiny{$\pm$0.48} & 78.13\tiny{$\pm$1.13} & \textbf{56.04}\tiny{$\pm$7.44} & \textbf{59.27}\tiny{$\pm$8.56} & \textbf{80.32}\tiny{$\pm$0.53} & \textbf{72.44}\tiny{$\pm$0.36} & \textbf{69.77}\tiny{$\pm$0.93} & \textbf{70.91}\tiny{$\pm$0.30} & \textbf{43.29}\tiny{$\pm$0.53} & \textbf{49.39}\tiny{$\pm$0.64} & \textbf{61.18} \\
\hline
\end{tabular}
% $}
\caption{Results (\%) against different ratios of label noises. RoBERTa is applied as the model backbone. 
}
  \label{exp:noise}
\end{table*}

\begin{figure*}[t]
\centering
\includegraphics[width=0.95\linewidth]{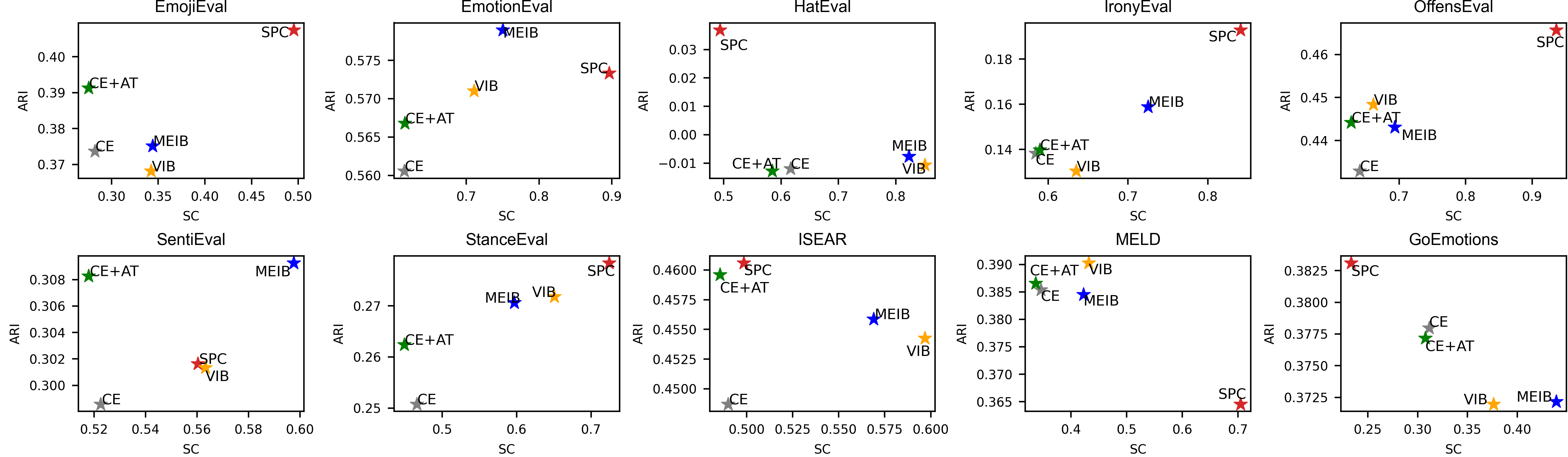}  
\caption{Clustering performances of the output representations learned by different optimization objectives. Silhouette coefficient (SC) and adjusted rand index (ARI) are used to measure data-related and task-related clustering abilities, respectively.
We experiment with RoBERTa backbone.}
\label{fig:struct}
\end{figure*}

\subsection{Robustness Evaluation}
We experiment to demonstrate the robustness by assessing how well the models can handle noisy labels. 
It is crucial for real-world scenarios where data can often be unreliable.
Specifically, we randomly choose 10\%, 20\%, 30\% of training data and flip their labels to any category randomly with equal probability. 
We run experiments five times and compute the mean and standard variance of the results. As shown in Table~\ref{exp:noise}, 
under all settings, SPC consistently outperforms CE, VIB, and MEIB. 
It indicates that SPC performs more robustly on noisy training data. 
Besides, compared to CE, SPC improves average performance by \textbf{+1.9\%}, \textbf{+2.2\%}, and \textbf{+1.6\%}  with noise ratio of 10\%, 20\%, and 30\% on classification tasks. 
The results prove that SPC can better control and utilize randomness and uncertainty of data.

\subsection{Representation Quality Evaluation}
To assess the quality of the representations, we evaluate the clustering performance of output representations obtained by different optimization objectives.
Following \citet{DBLP:conf/acl/0001BWZH23}, we apply silhouette coefficient (SC) and adjusted rand index (ARI) to measure the clustering ability relevant to input data and target labels, respectively.
Figure~\ref{fig:struct} shows SC and ARI of representations learned by various learning objectives. 
According to the results, SPC achieves higher ARI or SC values compared to other objectives (CE, VIB, and MEIB) across most datasets.
It suggests that SPC effectively achieves a balance between data encoding and task prediction, thereby promoting the generalization of pre-trained language models for downstream tasks.

\section{Related Work} 
Representation learning provides low-dimensional representations of the data, which can effectively capture the data features and serve for various tasks.
According to the nature of embeddings, it can be broadly categorized into deterministic embedding and probabilistic embedding.

\paragraph{Deterministic Embedding}
Deterministic embedding maps each data point to a fixed vector. The representative works include entropy regularization \citep{DBLP:conf/iclr/PereyraTCKH17}, adversarial training \citep{DBLP:conf/iclr/MiyatoDG17}, and contrastive learning \citep{gunel2020supervised,DBLP:conf/acl/0001BWZH23}.

\paragraph{Probabilistic Embedding}
Probabilistic embedding 
\cite{DBLP:journals/corr/VilnisM14} learns a probability distribution that maps each data point to a distribution. The probabilistic approach can better capture the complexity and uncertainty of data, handle redundant information, and provide better discriminative representations. It has been applied to various domains such as computer vision \cite{DBLP:conf/iclr/OhMPRSG19,DBLP:conf/iccv/ShiJ19,chang2020data} and natural language processing \cite{DBLP:conf/iclr/MahabadiBH21,DBLP:conf/iclr/WangWCGJLL21,DBLP:conf/emnlp/0001HDZJMS22}.

Most probabilistic embedding methods \cite{DBLP:journals/corr/KingmaW13,DBLP:conf/iclr/AlemiFD017,DBLP:conf/iclr/HigginsMPBGBML17,DBLP:journals/entropy/Fischer20,DBLP:conf/cvpr/AnJC23} are (or can be) built upon the principle of information bottleneck (IB) theory \cite{tishby1999information,tishby2015deep}.
The principle aims to find a maximally compressed representation of the input that maximally preserves information about the output, striking a balance between compression and prediction. 
VIB \cite{DBLP:conf/iclr/AlemiFD017} is an efficient variational estimation method of the IB principle. 
For tractable application of IB in a continuous setting, \citet{DBLP:conf/icml/BelghaziBROBHC18} propose a mutual information neural estimation method with IB principle, denoted as MINE-IB.
And \citet{DBLP:conf/cvpr/RagonesiVCM21} employ MINE-IB to learn unbiased representations.
\citet{DBLP:journals/entropy/Fischer20} and \citet{DBLP:conf/iclr/RameC21} introduce a conditional mutual information term to alleviate the over- or under-compression issue of traditional IBs.
Moreover, 
variational autoencoder (VAE) \cite{DBLP:journals/corr/KingmaW13} is a special case of an unsupervised VIB and can be used to encourage disentangled representations
\cite{DBLP:conf/iclr/HigginsMPBGBML17}.  
\citet{DBLP:conf/emnlp/0001HDZJMS22} 
apply VAE to the masked pre-training process for learning diverse and well-formed contextual representations.
Recently, \citet{DBLP:conf/cvpr/AnJC23} use the conditional entropy of the stochastic embedding as a confidence indicator and encourage the model to assign larger variance to more certain inputs.

\section{Conclusion}
This paper proposes a new supervised representation learning framework SPC, an encoder-only probabilistic coding technology with a structured regularization from the target space.
By extracting compact and informative representations from input related to the target task, SPC can enhance the generalization ability of pre-trained language models for better language understanding. 
Firstly, an encoder-only probabilistic coding technology simultaneously performs variational encoding and task prediction. 
Then, a structured regularization is introduced to control probability distribution and promote uniformity across classes in the latent space. 
Experiments on 12 benchmarks show that SPC achieves the best performance on various classification and regression tasks. The results demonstrate that SPC can enhance the generalization capability, robustness to label noise, and the clustering quality of output representations. 

\appendix

\section*{Appendix Overview}
In this supplementary material, we provide: 
(i) a detailed description of experimental setups,
and (ii) supplementary experiments.

\section{Experimental Setups}
\subsection{Datasets and Downstream Tasks} 

We conduct extensive experiments on various natural language understanding tasks including 10 classification  tasks, and 2 regression tasks, as shown in Table~\ref{tab:datasets}. The descriptions of each dataset and task are listed as follows:

\paragraph{Classification Tasks}
\begin{itemize}
\item 
\textbf{EmojiEval} \cite{DBLP:conf/semeval/BarbieriCRABBPS18} is designed for emoji prediction, which aims to predict its most likely emoji given a tweet. Its label set comprises 20 different emoji.  
\item 
\textbf{EmotionEval} \cite{DBLP:conf/semeval/MohammadBSK18} involves detecting the emotion evoked by a tweet and is based on the Affects in Tweets conducted during SemEval-2018.
Following \citet{DBLP:conf/emnlp/BarbieriCAN20}, the most common four emotions (i.e., anger, joy, sadness, and optimism) are selected as the label sets.
\item \textbf{HatEval} \cite{DBLP:conf/semeval/BasileBFNPPRS19} stems from SemEval-2019 Hateval challenge and is used to predict whether a tweet is hateful towards immigrants or women.
\item
\textbf{IronyEval} \cite{DBLP:conf/semeval/HeeLH18} is from SemEval-2018 Irony Detection and consists of identifying whether a tweet includes ironic intents or not.
    \item 
\textbf{OffensEval} \cite{DBLP:conf/semeval/ZampieriMNRFK19} is from SemEval-2019 OffensEval and involves predicting if a tweet contains any form of offensive language. 
    \item 
\textbf{SentiEval} \cite{DBLP:conf/semeval/RosenthalFN17} comes from SemEval 2017 and includes data from previous runs (2013,
2014, 2015, and 2016) of the same task. The goal is to determine if a tweet is positive, negative, or neutral. 
    \item 
\textbf{StanceEval} \cite{DBLP:conf/semeval/MohammadKSZC16} involves determining if the author of a piece of text has a favorable, neutral, or negative position towards a proposition or target. 
    \item 
\textbf{ISEAR} \cite{scherer1994evidence} is from International Survey On Emotion Antecedents And Reactions project and contains reports on seven emotions each by close to 3000 respondents in 37 countries on all 5 continents.
It aims to predict the emotion reaction.
Due to the lack of a predefined split in the original dataset, we randomly split the dataset into train/valid/test set in a ratio of 4:1:5 based on the label distribution.
    \item 
\textbf{MELD} \cite{DBLP:conf/acl/PoriaHMNCM19} contains multi-party conversation videos collected from Friends TV series, where two or more speakers are involved in a conversation.  It is used to detect emotions in each utterance.\footnote{The MELD dataset contains many types of context, including dialogue, speaker, and multi-modal data. Different from other task-oriented methods, e.g., DialogueCRN \cite{DBLP:conf/acl/HuWH20}, this work only considers the context-free textual utterance to better evaluate sentence classification performance.}
    \item 
\textbf{GoEmotions} \cite{DBLP:conf/acl/DemszkyMKCNR20} is a corpus of comments from Reddit, with human annotations to 27 emotion categories or neutral. It is used for fine-grained emotion detection.
In this work, we remove all multi-label samples (nearly 16\%) in the dataset to better evaluate the multi-class classification performance.
\end{itemize}

\begin{table*}[!h]
\centering  
\small
% \tabcolsep=1.5pt
% \resizebox{0.96\linewidth}{!}{$
\begin{tabular}{llrrrrr} 
  \hline
  \multirow{1}{*}{\textbf{Dataset}} &
    \multirow{1}{*}{\textbf{Task}} 
    & \multicolumn{1}{c}{\# \textbf{Label}}  
    & \multicolumn{1}{c}{\# \textbf{Train}} 
    & \multicolumn{1}{c}{\# \textbf{Val}}  
    & \multicolumn{1}{c}{\# \textbf{Test}} 
    & \multicolumn{1}{c}{\# \textbf{Total}} 
    \\ 
    \hline
    \multicolumn{3}{l}{\textit{Classification}} \\ 
EmojiEval     &  Emoji prediction            & 20  &   45,000  &   5,000   &   50,000 & 100,000 \\ 
EmotionEval  & Social emotion detection          & 4   &   3,257   &   374     &   1,421 &  5,052 \\ 
HatEval   & Hate speech detection             & 2   &   9,000   &   1,000   &   2,970 & 12,970 \\ 
IronyEval     &   Irony detection            & 2   &   2,862   &   955     &   784 & 4,601\\ 
OffensEval &  Offensive language detection          & 2   &   11,916  &   1,324   &   860 & 14,100  \\ 
SentiEval &  Sentiment analysis              & 3   &   45,389  &   2,000   &   11,906 & 59.295  \\ 
StanceEval    &  Stance detection            & 3   &   2,620   &   294     &   1,249  & 4,163  \\ 
ISEAR     &  Emotion reaction prediction      & 7   &   3,066   &    767    &   3,833  & 7,666 \\
MELD      &  Conversational emotion recognition       &  7   &   9,989    &   1,109  &   2,610 & 13,708\\
GoEmotions &  Fine-grained emotion detection    & 28  &   36,308  &   4,548   &   4,591 & 45,447  \\
\hline
    \multicolumn{3}{l}{\textit{Regression }} \\ 
STS-B &  Semantic similarity prediction & - & 7,000 & 1,500 & 1,400 & 9,900 \\
CLAIRE & Plausible clarification ranking & - & 19,975 & 2,500 & 2,500 & 24,975 \\ \hline 
  \end{tabular}
  % $}
  \caption{The statistics of all datasets. }
  \label{tab:datasets}
\end{table*}

\begin{table*}[!h]
\centering
\small
\tabcolsep=1.8pt
% \resizebox{0.96\linewidth}{!}{$
\begin{tabular}{l|cccccccccc}
\hline 
\multicolumn{1}{c|}{\multirow{1}{*}{\textbf{Hyperparameter}}} 
& \multicolumn{1}{c}{{EmojiEval}} 
& \multicolumn{1}{c}{{EmotionEval}} 
& \multicolumn{1}{c}{{HatEval}} 
& \multicolumn{1}{c}{{IronyEval}} 
& \multicolumn{1}{c}{{OffensEval}} 
& \multicolumn{1}{c}{{SentiEval}} 
& \multicolumn{1}{c}{{StanceEval}} 
& \multicolumn{1}{c}{{ISEAR}} 
& \multicolumn{1}{c}{{MELD}} 
& \multicolumn{1}{c}{{GoEmotions}}  \\ 
\hline 
Trade-off weight $\beta$  & 
1 & 0.1 & 10 & 1 & 0.01  &  0.01 & 1 & 0.1 & 1 & 0.001  \\
Trade-off weight $\gamma$  
& 10 & 1 & 0.1 & 0.1 & 0.1 &  10 & 0.1 & 0.1 & 0.1 & 0.001 \\
Weight decay & 0  & 0 & 0.01  & 0 & 0   &   0 & 0.001  & 0.001 & 0  & 0    \\ 
Dropout & 
0 & 0 & 0 & 0.2 & 0.2 & 0 & 0.2 & 0 & 0 & 0.2  \\
Layer normalization &  True & True & False & True & True &   False & True & False & True & False   \\
\hline 
\end{tabular}
% $}
\caption{Hyperparameters of the proposed SPC with RoBERTa backbone on classification tasks.}
\label{tab:spc}
\end{table*}
\begin{table}[!h]
\centering
\small
% \resizebox{0.6\linewidth}{!}{$
\begin{tabular}{l|cc}
\hline 
\multicolumn{1}{c|}{\multirow{1}{*}{\textbf{Hyperparameter}}} 
& \multicolumn{1}{c}{{STS-B}} 
& \multicolumn{1}{c}{{CLAIRE}}  \\ 
\hline 
Trade-off weight $\beta$  & 0.01 & 0.1 
\\ 
Weight decay & 0  & 0   \\ 
Dropout & 0 & 0 \\ 
Layer normalization  & False & False \\
\hline 
\end{tabular}
% $}
\caption{Hyperparameters of the proposed SPC with RoBERTa backbone on regression tasks.}
\label{tab:spc1}
\end{table}

\begin{figure}[!ht]
\centering
\includegraphics[width=0.98\linewidth]{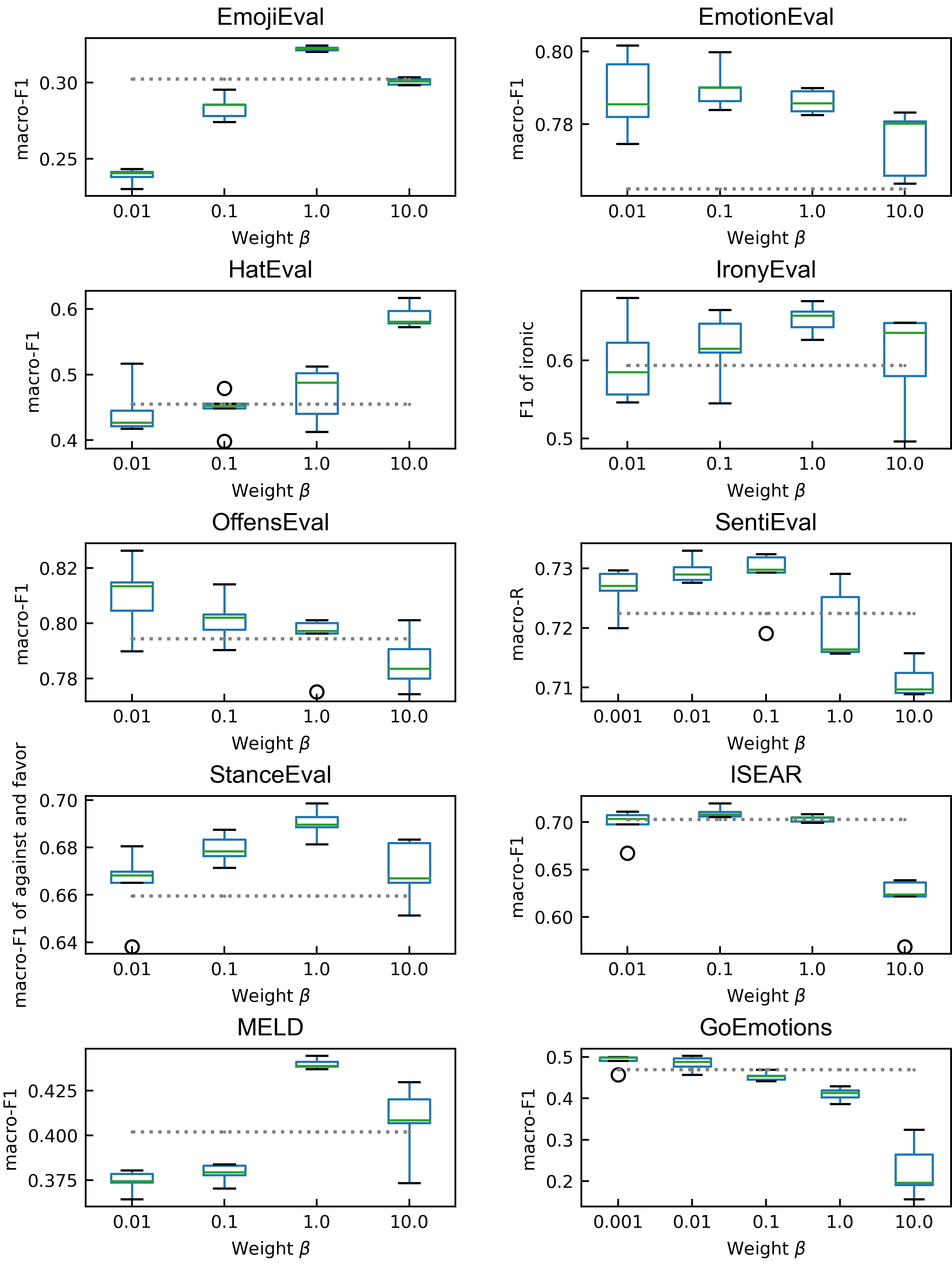}  
\caption{Performance against different trade-off weights $\beta$ of probabilistic coding for classification tasks. The experiments are conducted with RoBERTa backbone. The grey line represents the results of CE baseline.}
\label{fig:para}
\end{figure}
\begin{figure}[!ht]
\centering
\includegraphics[width=\linewidth]{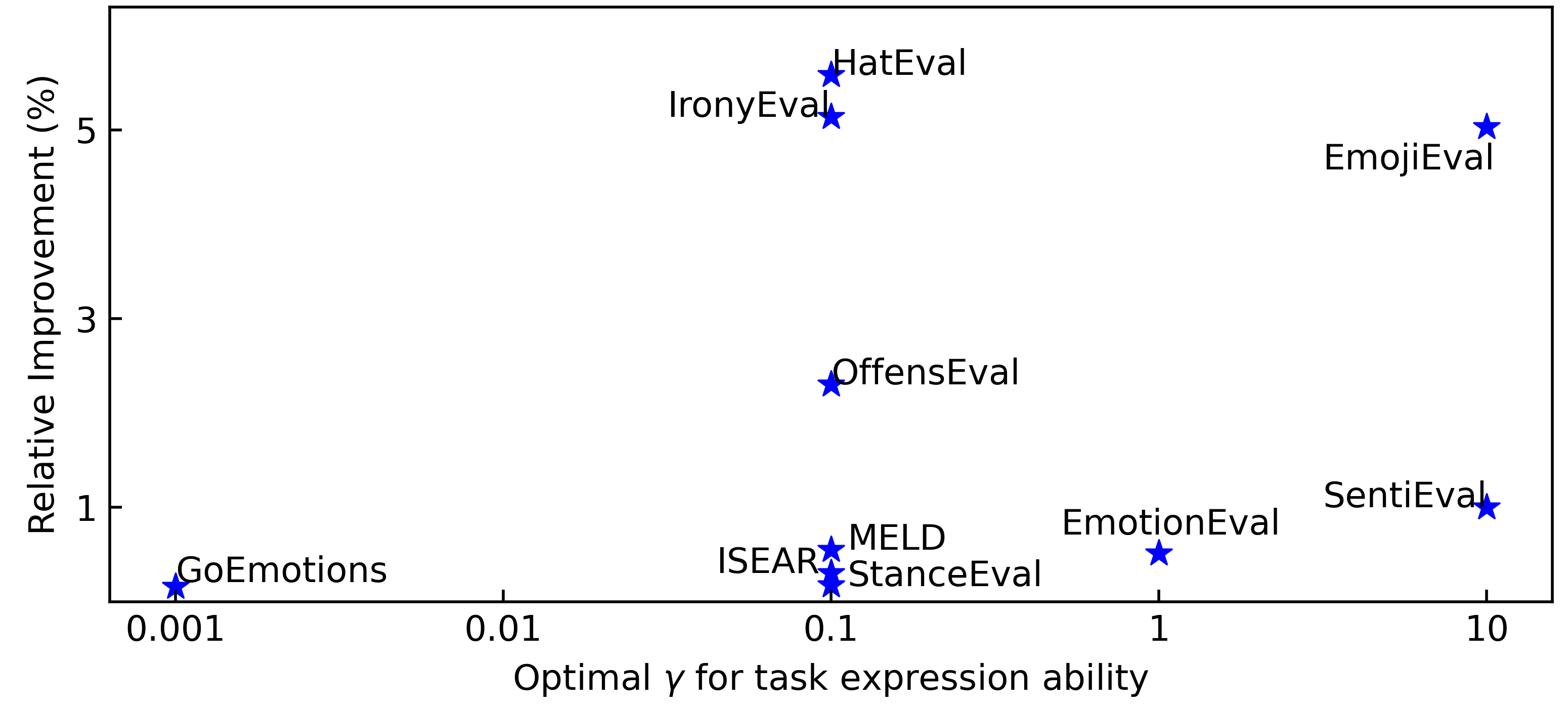}  
\caption{Performance of the optimal trade-off weight $\gamma$ for classification tasks. We experiment with RoBERTa backbone. The Y-axis refers to relative improvements between SPC and its variant removing the structured regularization. }
\label{fig:para_gam}
\end{figure}

\paragraph{Regression Tasks}
\begin{itemize}
    \item 
    \textbf{STS-B} \cite{DBLP:conf/semeval/CerDALS17} is a collection of English sentence pairs drawn from news headlines, video and image captions, and natural language inference data. The semantic similarity prediction task is to predict the semantic textual similarity score from 0 (very dissimilar) to 5 (very similar) given each sentence pair.
    \item   \textbf{CLAIRE} \cite{DBLP:conf/semeval/RothAS22} 
     dataset consists of manually clarified how-to guides from wikiHow\footnote{https://www.wikihow.com/} with generated alternative clarifications and human plausibility judgements.
    The goal of plausible clarifications ranking task is to predict the continuous plausibility score on a scale from 1 (very implausible) to 5 (very plausible) given the clarification and its context. 
    In our experiments, a special token pair (i.e., 
    \textless e\textgreater and \textless/e\textgreater
    ) is introduced as the boundary of filler words. % <e> and </e>
\end{itemize}

\subsection{Implementation Details}
We report the detailed hyperparameter settings of SPC with RoBERTa backbone in Table~\ref{tab:spc} and Table~\ref{tab:spc1}.
In the implementation of SPC, the hidden vector represents the output representation, and its dimension size is consistent with the dimension size of label space in each task. 

For each baseline, we fine-tune the key parameters following the original paper for fair comparison and to obtain corresponding optimal performance. In addition, the temperature of GPT-3.5 is set to 0 for deterministic predictions.

\section{Supplementary Experiments}
\subsection{Parameter Analysis}
In this part, we analyze what the trade-off weights $\beta$ and $\gamma$ control in our SPC.

% \paragraph{Performance against different weights of probabilistic coding}
Figure~\ref{fig:para} shows results against different values of $\beta$. With the enhancement of the optimization strength of probabilistic coding ($\beta \uparrow$), SPC is prone to assign larger variance for noisy samples and small variance for high quality ones.

% \paragraph{Performance against optimal weight of task expression}
Figure~\ref{fig:para_gam} shows relative improvements between SPC and its ablated variant (i.e., w/o Structured) against the optimal $\gamma$. By introducing the underlying structured patterns related to the target task ($\gamma>0$), SPC achieves varying degrees of relative improvements on all tasks, particularly in HatEval and IronyEval. A larger value of $\gamma$ indicates that this type of task requires enhanced task-related learning ability.

\section*{Acknowledgements} 
This work was supported by the National Key Research and Development Program of China (No. 2022YFC3302102), and the National Funded Postdoctoral Researcher Program of China (No. GZC20232969).
The authors thank the anonymous reviewers and the meta-reviewer for their helpful comments on the paper.

\bibliography{aaai24}

% \clearpage
\end{document}